\documentclass{article}

\usepackage[preprint]{neurips_2025}
\usepackage{caption}
\usepackage{float}

\usepackage[utf8]{inputenc} 
\usepackage[T1]{fontenc}    
\usepackage{hyperref}       
\usepackage{url}            
\usepackage{booktabs}       
\usepackage{amsfonts}       
\usepackage{algorithm}

\usepackage{algpseudocode}

\usepackage{amsmath}        
\usepackage{nicefrac}       
\usepackage{microtype}      
\usepackage{xcolor}         
\usepackage{natbib} 

\usepackage{multirow}  
\usepackage{makecell}  
\usepackage{colortbl}
\usepackage{booktabs}  
\usepackage{colortbl}  
\usepackage{multirow}  
\usepackage{graphicx}

\usepackage[bb=dsserif]{mathalpha}
\usepackage{bm}

\usepackage[utf8]{inputenc} 
\usepackage[T1]{fontenc}    
\usepackage{hyperref}       
\usepackage{url}            
\usepackage{booktabs}       
\usepackage{amsfonts}       
\usepackage{graphicx}       
\usepackage{nicefrac}       
\usepackage{microtype}      
\usepackage{xcolor}         
\usepackage{natbib} 
\usepackage{algorithm}
\usepackage{rotating}
\usepackage{multirow}  
\usepackage{makecell}  
\usepackage{amsmath}
\usepackage{amssymb}
\usepackage{mathtools}
\usepackage{amsthm}

\usepackage[capitalize,noabbrev,nameinlink]{cleveref}

\theoremstyle{plain}

\theoremstyle{definition}

\theoremstyle{remark}

\usepackage[textsize=tiny]{todonotes}
\usepackage{wrapfig}
\usepackage{microtype}
\usepackage{graphicx}
\usepackage{multirow}
\usepackage{subfigure}
\usepackage{booktabs} 
\usepackage{verbatim}
\usepackage{rotating}
\usepackage{hyperref}

\usepackage[capitalize,noabbrev,nameinlink]{cleveref}
\definecolor{indigo}{rgb}{0.29, 0.0, 0.51}   
\definecolor{amberyellow}{rgb}{0.85, 0.65, 0.13} 
\definecolor{rosepink}{rgb}{0.96, 0.52, 0.76}       
\definecolor{goldenrod}{rgb}{0.98, 0.84, 0.33}      
\definecolor{bluishpurple}{rgb}{0.61, 0.43, 0.86}   
\definecolor{royalpurple}{rgb}{0.47, 0.25, 0.74} 
\hypersetup{
    colorlinks=true,
    linkcolor=indigo,        
    filecolor=rosepink,            
    urlcolor=amberyellow,            
    citecolor=indigo,        
    pdftitle={Fine, I'll Merge It Myself: A Multi-Fidelity Framework for Automated Model Merging},
}

\usepackage{booktabs}  
\usepackage{colortbl}  
\usepackage{multirow}  
\title{Fine, I'll Merge It Myself: A Multi-Fidelity Framework for Automated Model Merging}

\author{
 Guinan Su\textsuperscript{1}
  \And
  Jonas Geiping\textsuperscript{1,2,3}
  \AND
  \textsuperscript{1}\normalfont Max Planck Institute for Intelligent Systems\\
  \textsuperscript{2}\normalfont ELLIS Institute Tübingen, 
  \textsuperscript{3}Tübingen AI Center
}

\begin{document}

\maketitle

\vspace{-0.3cm}
\begin{abstract}
\looseness -1 Large language models (LLMs) have steadily advanced in capability, but their development requires extensive proprietary datasets and computational resources. 
One way to efficiently supplement capabilities with is model merging, which offers a promising alternative by combining multiple models without retraining.  
However, current merging approaches rely on manually-designed strategies for merging hyperparameters, limiting the exploration of potential model combinations and requiring significant human effort. We propose an Automated Model Merging Framework that enables fine-grained exploration of merging strategies while reducing costs through multi-fidelity approximations. We support both single and multi-objective optimization and introduce two novel search spaces: layer-wise fusion (LFS) and depth-wise integration (DIS). 
Evaluating across a number of benchmarks, we find that the search autonomously finds 1) Merges that further boost single-objective performance, even on tasks the model has already been finetuned on, and 2) Merges that optimize multi-objective frontiers across tasks. Effective merges are found with limited compute, e.g. within less than 500 search steps. The code is available at.\footnote{\url{https://github.com/Guinan-Su/auto-merge-llm}}
\end{abstract}

\vspace{-0.3cm}
\section{Introduction}
\vspace{-0.2cm}

Recent advancements in large-scale pre-trained models like GPT-4 \citep{achiam2023gpt}, LLaMA \citep{touvron2023llama}, DALL·E \citep{ramesh2021zero}, and Imagen \citep{saharia2022photorealistic} have demonstrated remarkable capabilities across diverse domains. Within Large Language Models (LLMs), specialized models have emerged excelling in specific tasks such as instruction following \citep{xu2023wizardlm,jiang2023mistral,bi2024deepseek}, code generation \citep{luo2023wizardcoder,guo2024deepseek,zhu2024deepseek}, and mathematical problem-solving \citep{luo2023wizardmath}.
However, developing models with comprehensive capabilities remains challenging. A straightforward solution would be to combine training data from specialized domains for retraining or finetuning. However, this data-centric approach faces practical limitations: it requires substantial resources, and many training datasets remain proprietary or restricted. Researchers have turned to model-centric approaches that enhance capabilities by leveraging existing pre-trained models without  training or data access.

Model merging has emerged as a promising solution in this space. Beginning with simple weight averaging for models sharing initialization \citep{utans1996weight}, more advanced parameter-based techniques were subsequently developed. Parameter-based methods like Task Arithmetic \citep{ilharco2022editing} and SLERP \citep{white2016sampling} advanced the field through parameter difference computation and spherical interpolation. Recent sparsity-based approaches like TIES-Merging \citep{yadav2024ties} and DARE \citep{yu2024language} leverage neural network over-parametrization, using magnitude-based selection and rescaling to further improve merging effectiveness.
However, these model merging methods rely on manually-tuned hyperparameters, applied uniformly across the entire model, which makes finding optimal solutions challenging. 

In this paper, we propose a \textbf{cost-efficient automated framework} for model merging.  Our approach leverages low-fidelity approximations to reduce computational cost \citep{peherstorfer2018survey}  while supporting both single and multi-objective optimization. We further introduce layer-wise and depth-wise search spaces for finer-grained merging while also enhancing search efficiency by narrowing the search dimensions. Our key contributions include:

\begin{itemize}
\item We propose an automated model merging framework that enhances both single and multi-objective reasoning capabilities while reducing computational costs through multi-level fidelity optimization.
\item We introduce two novel search spaces: Layer-wise Fusion Space (LFS) for fine-grained layer merging and Depth-wise Integration Space (DIS) for optimizing inference pathways, enabling comprehensive model integration strategies while reducing search complexity.
\item We demonstrated efficient optimization across different scenarios: in mathematical reasoning using LFS, only 17\% of trials required the full search budget within 500 trials; in general reasoning using DIS, only 18.6\% of trials required the full budget within 1000 trials.

\item We achieve significant performance gains, including a 6.86\% average improvement in multi-objective scenarios and a 4.24\% improvement on the challenging GSM8K task, with consistent effectiveness across various reasoning benchmarks.
\end{itemize}

\vspace{-0.1cm}
\section{Related Work}
\vspace{-0.1cm}
\textbf{Model Merging} Model merging enhances capabilities without additional training data or extensive computation. The field evolved from simple weighted parameter averaging \citep{utans1996weight} to advanced methods. Task Arithmetic \citep{ilharco2022editing} computes parameter differences, SLERP \citep{white2016sampling} implements spherical interpolation, TIES-Merging \citep{yadav2024ties} selectively retains parameters based on magnitude while addressing sign conflicts, DARE \citep{yu2024language} combines magnitude-based sparsification with parameter rescaling. Recent Evolutionary model merging \citep{akiba2024evolutionary} optimizes coefficients through evolutionary search but is limited by uniform layer merging and excessive search space volume. Our framework introduces a fine-grained search space with reduced dimensionality, leveraging Multi-fidelity optimization \citep{peherstorfer2018survey} for efficient merging recipe optimization.

\textbf{Hyperparameter Optimization} Bayesian Optimization shows success in various applications \citep{snoek2012practical, mendoza2016towards}. Gaussian processes provide strong uncertainty estimates but lack scalability. Alternative models like Random forests \citep{hutter2011sequential} and Bayesian neural networks \citep{snoek2015scalable} address this limitation by offering better high-dimensional scaling. Hyperband, a multi-fidelity method \citep{peherstorfer2018survey}, allocates resources efficiently with successive halving \citep{jamieson2016non}, but its random sampling ignores previous evaluations. Our optimizer leverages SMAC \citep{JMLR:v23:21-0888}, combining Hyperband with Bayesian Optimization for efficient resource allocation and improved learning through surrogate modeling.

\section{Motivation}

 \begin{wrapfigure}{r}{0.55\textwidth}
  \centering
  \vspace{-1.5cm}
  \includegraphics[width=1\linewidth]{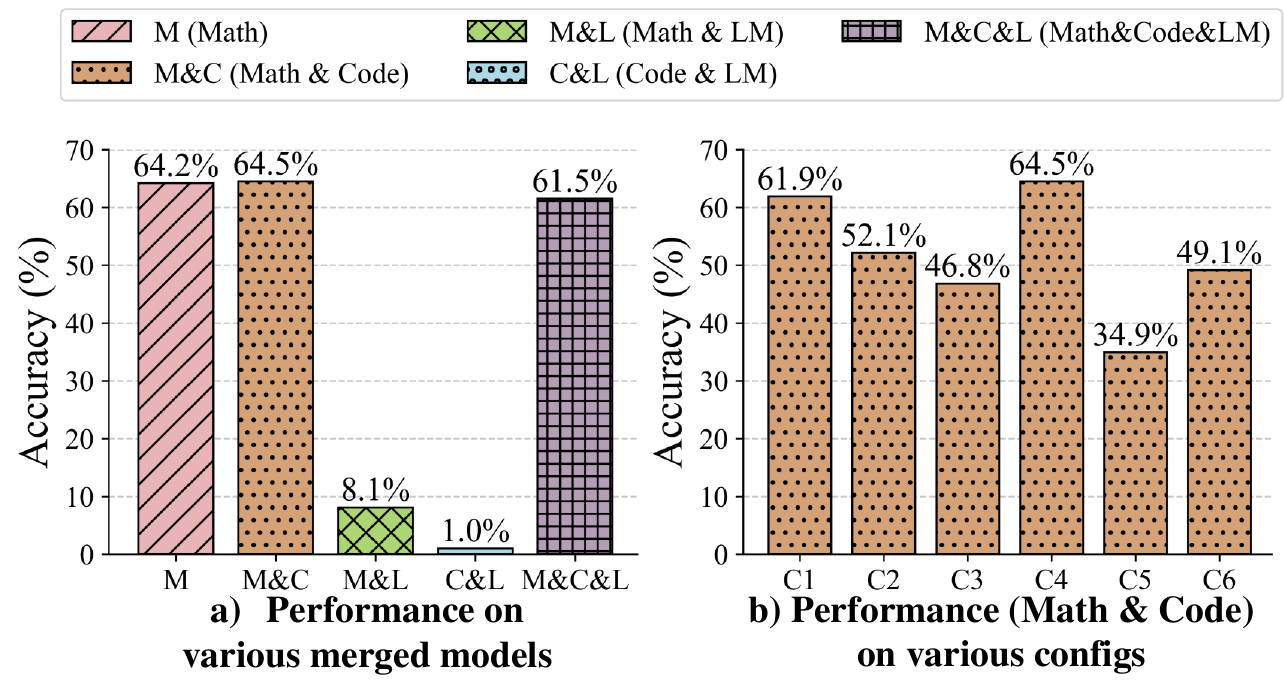}
  \caption{Performance evaluation with TIES on the GSM8K benchmark: (a) comparison of different source model combinations and (b) various configurations with Math and Code as source models.}
  \label{fig:fig2}
  \vspace{-0.3cm}
\end{wrapfigure}

Current merging methods typically incorporate layerwise and depthwise modifications to maintain architectural stability while avoiding costly post-finetuning \citep{ilharco2022editing,white2016sampling,kim2023solar}. However, despite their effectiveness, these approaches face inevitable limitations.

\textbf{Limitations of Current Approaches} For layer-wise merging, most methods~\citep{ilharco2022editing, white2016sampling, yadav2024ties, yu2024language} focus on combining corresponding layers from multiple models. While effective, these approaches typically apply uniform merging methods and hyperparameters across all layers, using the entire set of candidate models, which might be too coarse-grained and potentially problematic. 
To illustrate this concern, we conduct an analysis using TIES~\citep{yadav2024ties}, one of the most robust merging methods, combining WizardLM-13B(LM) \citep{xu2023wizardlm}, WizardMath-13B(Math) \citep{luo2023wizardmath} , and llama-2-13b-code-alpaca(Code) \citep{chaudhary2023code}  on the GSM8K \citep{cobbe2021training} benchmark. We set the scaling term and parameter retention ratio in reasonable ranges of [0.5, 1.0] and [0.5, 0.7, 0.9], respectively. See Table~\ref{tab:ties_performance} for detailed configuration information. As shown in Figure~\ref{fig:fig2}(a), when applying TIES merge with the same hyperparameters across different model combinations, the accuracy varies dramatically from 1\% to 64.5\%. Furthermore, Figure~\ref{fig:fig2}(b) demonstrates that even with a fixed model combination (Math+Code), different hyperparameter settings lead to substantial performance variations, ranging from 34.9\% to 64.5\%. These results reveal two critical challenges in layer-wise merging: the selection of candidate models and the determination of hyperparameters. Both factors significantly impact the final performance, even for well-established methods like TIES. For depth-wise merging approaches, such as Solar~\citep{kim2023solar}, which simply scales large language models through concatenation of models, the resulting models exhibit insufficient performance and require further expensive pretraining to meet downstream task requirements, which further underscores the critical importance of hyperparameter selection.  Manual tuning of these hyperparameters is not only labor-intensive but also makes it challenging to find optimal configurations.


\textbf{Automatic Model Merging} The above concerns highlight that automatic model merging remains largely unexplored. Recent advances include Evolutionary model merging~\citep{akiba2024evolutionary}, which optimizes merging coefficients through automatic evolutionary search. However,  It’s limited by uniform layer merging and exploration of all possible layer interactions, creating an excessively large search space that hinders optimization.  We aim to provide a more fine-grained and effective automatic merging framework. A comprehensive comparison with Evolutionary model merging~\citep{akiba2024evolutionary} appears in the Appendix \ref{sec:compare}. 
We present our methods in the following sections.





\begin{figure}
    \centering
    \vspace{-0.3cm}
    \includegraphics[width=0.9\textwidth]{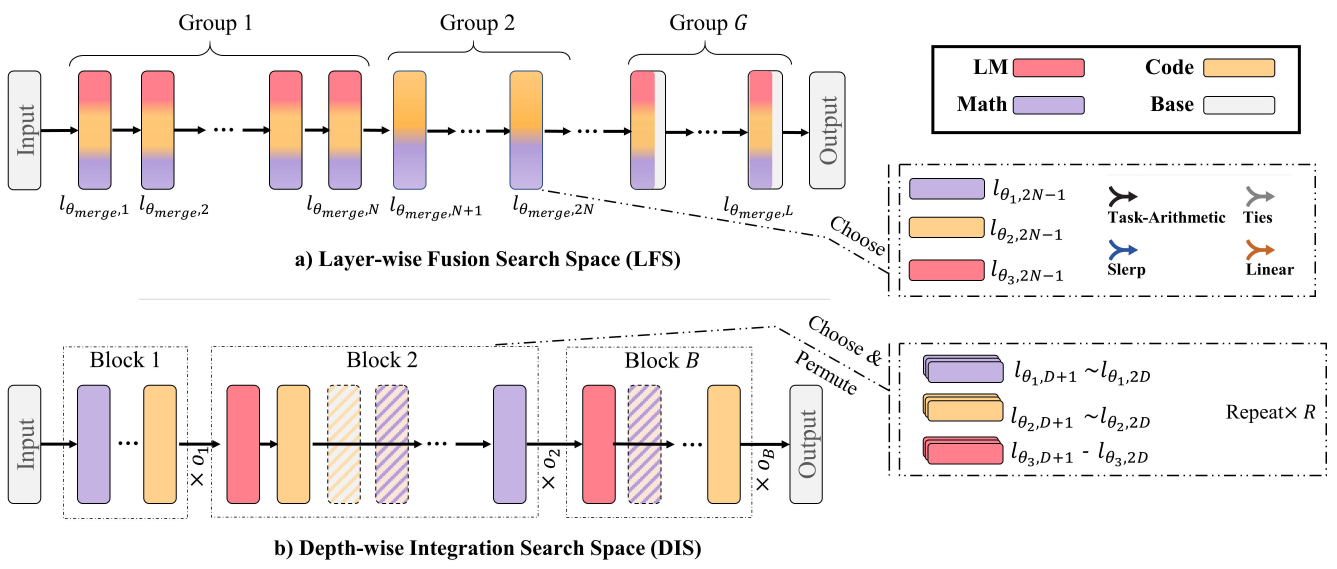}
    \caption{Illustrations of two merging search spaces: (a) Layer-wise Fusion Search (LFS), which merges layers across layer groups with different methods and hyper-parameters, and (b) Depth-wise Integration Search (DIS), which optimizes inference paths through block-wise searching with permutation and layer selection from different source models.}
    \label{fig:fig1}
  \vspace{-0.6cm}
\end{figure}
\section{Methods}

\subsection{Overview}

To define a hyperparameter optimization pipeline for model merging, we need three parts, 
a search space that determines how models are merged by designed configuration, objective functions that evaluate desired capabilities of the merged model, and an efficient search optimizer that identifies optimal configurations. \textbf{(1) Search spaces:} We introduce two complementary search spaces. LFS enables fine-grained merging of corresponding layers through optimal merge operations, while DIS addresses models lacking mergeable corresponding layers by optimizing sequential arrangement of preserved layer weights. \textbf{(2) Objectives:} Our framework supports both task-specific optimization and Pareto-optimal configurations across multiple reasoning dimensions. \textbf{(3) Optimizer:} We implement a multi-fidelity optimizer that accelerates search by efficiently allocating computational resources to promising candidate architectures.

\subsection{Search Space}

\subsubsection{Layer-wise Fusion Search Space}


We design a fine-grained layer-wise merging search space (LFS). Our search space is illustrated in Figure \ref{fig:fig1} (a), we partition the model's $L$ layers into $G$ consecutive groups, where layers within each group share the same merging coefficients. These coefficients determine: (1) the selection of source models from the candidate pool, (2) the choice of merging algorithms, and (3) the corresponding hyperparameters for the chosen merging method.
Inspired by studies showing specialized functions in neural architectures where MLPs store facts \citep{geva2020transformer} and attention captures relations between concepts \citep{wang2022interpretability}, we further introduce a component-wise decomposition strategy. Specifically, we partition the parameters within each Transformer layer into $C$  component groups. When $C=1$, the entire layer is treated as a single unit. When $C=3$, we decompose the layer into three groups: MLP-related parameters, attention mechanism parameters, and layer normalization parameters. This decomposition facilitates a better way of preserving the unique functional contributions of each architectural element.

We define the merging coefficients $x \in \mathbb{R}^{G \times C \times (1 + H)}$, where $G$ represents the number of layer groups, $C$ denotes the number of components per layer, and $1 + H$ dimensions specify the merging method selection and hyperparameters of all candidate merging methods. We use four well-established merging methods: Task Arithmetic, TIES-Merging, SLERP, and Linear Merging. See Section \ref{sec:a0} for additional information. LFS provides a fine-grained and flexible search space for model merging through multiple merging methods, layer selection, and granular controls, which not only enables precise optimization of the fusion but also maximizes the potential of layer-wise merging.


\subsubsection{Depth-wise Integration Search Space}

Large Language Models (LLMs) exhibit hierarchical language understanding, with knowledge transformation progressing sequentially from word-level comprehension to abstract conceptual understanding. Recent research has increasingly focused on the behavior of transformer layers. \citet{meng2022locating} and \citet{geva2022transformer} explored the distribution of knowledge across different layers.  \citet{lad2406remarkable} examined the robustness of transformer-based large language models by analyzing the effects of layer deletion and swapping. \citet{sun2024transformer} show that lower and final layers differ from middle layers, which demonstrate uniformity and robustness to reordering and parallelization for certain tasks. This hierarchical relationship between layers remain unexplored in the context of model merging. To fill this gap, we introduce the Depth-wise Integration Search Space (DIS), which preserves the original weights of individual layers while optimizing the inference pathway.


As depicted in Figure \ref{fig:fig1} (b), DIS is characterized by three parameters: depth granularity $D$, number of candidate models $M$, and repeat factor $R$. These parameters partition transformer layers into $B = L/D$ consecutive blocks, where each block encompasses $D \times M \times R$ candidate layers. The search space is characterized by merging coefficients $x = \{(\mathbf{s}^{(i)}, \mathbf{p}^{(i)}, o_i)\}_{i=1}^B$, where selection vector $\mathbf{s}^{(i)} \in \{0,1\}^{M \times D \times R}$ determines layer activation and permutation vector $\mathbf{p}^{(i)} \in \{0, 1, ..., P-1\}$ with $P = (D \times M \times R)!/(R!)^{M \times D}$ specifies layer ordering, and scaling factor $o_i$ normalizes the output of each block \citep{akiba2024evolutionary}. When no layers are selected in a block ($\mathbf{s}^{(i)} = 0$), we implement a layer retention strategy that preserves model depth by defaulting to base model layers.

The parameterization enables various architectural operations (pruning, stacking, repetition, reordering) controlled by granularity parameter $D$. At $D=1$, the search space focuses on position-corresponding layer interactions, while increasing $D$ allows more complex integration patterns and cross-depth interactions. This approach controls layer interaction through depth granularity, enabling flexible task adjustment while reducing search space with fewer hyperparameters.




\subsection{Objectives}

\textbf{Single objective} We use single-objective to maximize task-specific performance. The cost function for each task is defined as $
c_i(x) = \arg\min_{x \in \Lambda} \mathcal{L}(\mathcal{D}_i; x)$.
where $c_i(x)$ represents an optimization objective over the parameter space $x$ for a specific task dataset $\mathcal{D}_i$.  The loss function $\mathcal{L}$ measures the model's performance on the target task dataset.

\textbf{Multi objective} 
To develop comprehensive reasoning models, we employ ParEGO \citep{knowles2006parego} for multi-objective optimization to identify Pareto-optimal solutions across different objectives. The algorithm converts multiple cost functions into a single aggregated cost using a parameterized scalarizing weight vector. By varying this weight vector at each iteration, ParEGO gradually builds an approximation of the entire Pareto front. Initially, the algorithm normalizes the $k$ cost functions $c_j(x)$ to the $[0,1]$ interval. In each step, the algorithm randomly selects a weight vector $\boldsymbol{\lambda}$ from a set of uniformly distributed vectors, defined as:

\begin{equation*}
\begin{aligned}
 \Lambda=\biggl\lbrace\boldsymbol{\lambda}=\left(\lambda_1, \lambda_2, \ldots, \lambda_k\right) \mid & \sum_{j=1}^k \lambda_j=1 \wedge \forall j,
  \lambda_j=\frac{l}{s}, l \in\{0, \ldots, s\}\biggr\rbrace
\end{aligned}
\end{equation*}

The size of this set is determined by $|\Lambda| = \binom{s+k-1}{k-1}$, where $s$ controls the total number of possible vectors. The aggregated cost $c_{\text{agg}}$ for each solution $x$ is calculated using the augmented Tchebycheff function, where $\rho$ is a small positive constant that ensures Pareto optimality:

\begin{equation}
c_{\text{agg}}(x;\boldsymbol{\lambda}) = \max_{j=1}^k(\lambda_j \cdot c_j(x)) + \rho\sum_{j=1}^k \lambda_j \cdot c_j(x) 
\end{equation}
\vspace{-0.3cm}


\subsection{Multi-Fidelity Optimization}
Although optimization of model merging requires less computation compared to searching for optimal hyper-parameters for neural network structures and the model training process~\citep{yu2020hyper,elsken2019neural}, evaluating large language models on extensive validation datasets remains computationally intensive. We optimize the process using cost-efficient Multi-Fidelity Optimization (MFO)~\citep{peherstorfer2018survey}, leveraging evaluations across different fidelity levels from fast but less accurate low-fidelity to slow but more accurate high-fidelity. 
The optimization objective can be formulated as $
x^* \in \arg\min_{x \in \Lambda} c(x, b)$.
Here, we use evaluation sample size as fidelity types, represented by budgets $b$ where $b_{\text{min}} \leq b \leq b_{\text{max}}$. Each configuration is evaluated with varying budgets $b_{\text{min}} \leq b \leq b_{\text{max}}$, Using smaller budgets provides a cheaper proxy of the true cost function.



Our implementation extends SMAC~\citep{JMLR:v23:21-0888} by establishing a hierarchical evaluation framework parameterized by $b_{\text{max}}$, $b_{\text{min}}$, and spacing factor $\eta$. The number of brackets is determined by $s_{\text{max}} = \lfloor \log_{\eta}(b_{\text{max}}/b_{\text{min}}) \rfloor$, with each bracket $s$ starting with $n_s = \lceil \frac{s_{\text{max}} + 1}{s + 1} \cdot n \rceil$ configurations at budget $b_s = b_{\text{min}} \cdot \eta^s$. Using Successive Halving~\citep{jamieson2016non}, each bracket iteratively halves configurations and increases budgets by $\eta$ until reaching $b_{\text{max}}$. A Random Forest surrogate model~\citep{breiman2001random} guides configuration selection through Expected Improvement. Optimization terminates when total budget is exhausted, maximum iterations reached, or performance plateaus. See Section \ref{sec:a1} for more descriptions of the optimization.

\section{Experiments}
\subsection{Experimental Setting}

\textbf{Source Models}
We use Llama-family models \citep{touvron2023llama} as base model set, including WizardLM-13B(LM) \citep{xu2023wizardlm}, WizardMath-13B(MATH) \citep{luo2023wizardmath}, and llama-2-13b-code-alpaca(CODE) \citep{chaudhary2023code}. All these models are fine-tuned from Llama-2-13b, ensuring a shared loss landscape. We exclude WizardCoder-Python-13B~\citep{luo2023wizardcoder} as it uses a different pre-trained backbone, resulting in a different loss landscape.

\textbf{Datasets}
We select separate datasets for search and evaluation. For searching, we use GSMPlus \citep{li2024gsm} for mathematical reasoning, MBPP \citep{austin2021program} samples for code understanding, and MMLU \citep {hendrycks2020measuring}validation samples for general knowledge. For evaluation, we employ established benchmark test sets: GSM8K \citep{cobbe2021training} and MATH \citep{hendrycksmath2021} for mathematical reasoning, MBPP and HumanEval \citep{chen2021codex} for code generation, and the MMLU test set for general knowledge assessment. See Section \ref{sec:a2} and \ref{sec:a3} for more details.


\textbf{Search Spaces} 
\looseness -1 As described in the method section, within LFS we implement four merging methods: Task Arithmetic, TIES-Merging, Linear-merging, and Slerp. We set the number of layer groups to $G=4$ and the number of component groups per layer to $C=3$. For DIS, we evaluate two configurations: Configuration $1$ with depth granularity $D=1$, number of candidate models $M=3$, and repeat factor $R=1$; Configuration $2$ with $D=1$, $M=1$, and $R=2$. Both configurations keep $D=1$ because reasoning tasks exhibit sensitivity to layer ordering, where variations can degrade performance \cite{sun2024transformer}, constraining broader exploration. We focus on layer-wise interactions between corresponding positions across candidate models. The two configurations were designed to be complementary: Configuration $1$  targets interactions across models, whereas Configuration $2$ targets repeated layers within a model. We set a maximum layer constraint of $50$ to manage computational costs.

\textbf{Objectives and Optimization} 
We define task-specific single objectives using accuracy performance on specialized datasets: GSMPlus \citep{li2024gsm} for mathematical reasoning, MBPP \citep{austin2021program} validation set for programming capabilities, and MMLU \citep{hendrycks2020measuring} validation set for general reasoning. For multi-objective optimization, we seek Pareto-optimal solutions across these three tasks.
Our optimizer builds upon SMAC~\citep{JMLR:v23:21-0888}, with domain-specific budget allocations for optimization tasks. Mathematical reasoning tasks receive 100-1000 samples to explore complex solution spaces, code reasoning uses 300 samples (200 training, 100 validation), and we sample 50\% of MMLU validation data within 100-700 sample bounds for general reasoning. These budget ranges remain consistent when conducting multi-objective optimization across multiple datasets within these domains. We configured the search trials based on the complexity of each search space. We allocated $500$ search trials for LFS and $1000$ for the broader DIS search space, using initial candidate models as starting points to improve optimization efficiency.




\subsection{Results}

\begin{table*}[t]
\centering
\scriptsize
\setlength{\tabcolsep}{1pt}
\caption{Performance comparison of merged models combining WizardLM-13B (LM), WizardMath-13B (Math), and llama-2-13b-codealpaca (Code). For single-objective optimization, improvements over source models are shown in blue. For multi-objective optimization, improvements over the best base model (Math) are shown in blue in the Average column.}
\begin{tabular}{@{}c|c|c|c|lllll|c@{}}
\toprule
& & & & \multicolumn{1}{c}{\textbf{Common}} & \multicolumn{2}{c}{\textbf{Math}} & \multicolumn{2}{c}{\textbf{Code}} & \\
\cmidrule(lr){5-5} \cmidrule(lr){6-7} \cmidrule(lr){8-9}
\textbf{Method} & \textbf{Model} & \textbf{Source\textsuperscript{*}} & \textbf{Search\textsuperscript{*}} & MMLU & GSM8K & MATH & MBPP & HumanEval & Average \\
\midrule
\multirow{3}{*}{\textbf{Base}} 
& Math & -- & -- & 52.04 & 64.22 & \textbf{13.70} & 18.20 & 7.32 & 31.10 \\
& Code & -- & -- & 52.79 & 0.00 & 0.00 & 27.20 & 23.17 & 20.63 \\
& LM & -- & -- & 53.43 & 3.79 & 0.00 & \textbf{33.40} & \textbf{38.41} & 25.81 \\
\midrule
\multirow{3}{*}{\textbf{Basic}} 
& Ties & M+C+L & -- & 54.67 & 61.56 & 10.58 & 27.40 & 23.17 & 35.48 \textcolor{blue}{(+4.38)} \\
& Task Arith & M+C+L & -- & 54.85 & 57.99 & \textbf{12.06} & 26.22 & \textbf{24.04} & 35.03 \textcolor{blue}{(+3.93)} \\
& Linear & M+C+L & -- & \textbf{55.13} & 57.09 & 9.98 & 29.80 & 18.90 & 34.18 \textcolor{blue}{(+3.08)} \\
\midrule
\multirow{5}{*}{\textbf{Evo-Single-Obj}} 
& MATH-LFS-EVO & M+C+L & 1 & 54.46 & \underline{63.53} \textcolor{blue}{(-0.69)} & 10.76 & 30.80 & 17.07 & -- \\
& CODE-LFS-EVO & M+C+L & 2 & 54.99 & 64.29 & 11.16 & \underline{31.00} \textcolor{blue}{(-2.40)} & 23.17 & -- \\
& GEN-LFS-EVO & M+C+L & 3 & \underline{55.00} \textcolor{blue}{(+1.57)}& 51.25 & 8.58 & 30.40 & 25.61 & -- \\
\midrule
\multirow{5}{*}{\textbf{Ours-Single-Obj}} 
& MATH-LFS & M+C+L & 1 & 54.52 & \underline{\textbf{68.46}} \textcolor{blue}{(+4.24)} & 10.42 & 28.20 & 17.07 & -- \\
& CODE-LFS & M+C+L & 2 & 53.36 & 49.73 & 9.30 & \underline{\textbf{33.60}} \textcolor{blue}{(+0.20)} & 14.63 & -- \\
& GEN-LFS & M+C+L & 3 & \underline{\textbf{55.31}} \textcolor{blue}{(+1.88)}& 33.81 & 3.82 & 30.80 & 16.46 & -- \\
& GEN-DIS-0 & M+C+L & 3 & \underline{54.72} \textcolor{blue}{(+1.29)}& 16.98 & 1.14 & 12.40 & 9.76 & --  \\
& GEN-DIS-1 & L & 3 & \underline{54.76} \textcolor{blue}{(+1.33)} & 1.74 & 0.06 & 16.80 & 18.29 & -- \\
\midrule
\multirow{5}{*}{\textbf{Ours-Multi-Obj}} 
& MULTI-LFS-0 & M+C+L & 1-3 & \textbf{55.03} & 63.08 & 11.76 & \textbf{32.60} & 21.95 & 36.88 \textcolor{blue}{(+5.78)} \\
& MULTI-LFS-1 & M+C+L & 1-3 & 54.70 & \textbf{66.94} & 11.38 & 30.60 & 23.78 & 37.48 \textcolor{blue}{(+6.38)} \\
& MULTI-LFS-2 & M+C+L & 1-3 & 54.67 & 65.13 & 11.06 & 30.40 & 20.73 & 36.40 \textcolor{blue}{(+5.30)} \\
& MULTI-LFS-3 & M+C+L & 1-3 & 54.99 & 65.50 & 9.42 & 30.20 & 23.17 & 36.66 \textcolor{blue}{(+5.56)} \\
& MULTI-LFS-4 & M+C+L & 1-3 & 54.77 & \textbf{66.79} & \textbf{12.06} & 31.80 & \textbf{24.40} & 37.96 \textcolor{blue}{(+6.86)} \\
\bottomrule 
\multicolumn{10}{l}{\textsuperscript{*}Dataset Index: 1=GSMPlus, 2=MBPP\textsubscript{val}, 3=MMLU\textsubscript{val}}\\
\multicolumn{10}{l}{\textsuperscript{*}M+C+L = Math+Code+LM}\\
\end{tabular}
\label{tab:table0}
\vspace{-.1cm}
\end{table*}
\subsubsection{Single objective optimization}

\textbf{Layer-wise Fusion} Using three source models (Math, Code, and LM) optimized for mathematical, coding, and general reasoning tasks respectively, we developed three specialized models through LFS: MATH-LFS, CODE-LFS, and GEN-LFS.  Table~\ref{tab:table0} presents the performance of these models across five benchmarks. Our results demonstrate that MATH-LFS achieves a 4.24\% improvement over the best performance of source models on GSM8K, CODE-LFS shows modest gains on MBPP, and GEN-LFS exhibits a 1.88\% improvement on MMLU. These MATH-LFS gains are especially surprising, given that the base model was already finetuned for improve GSM8K performance - it appears that its arithmetic performance could be further improved through merging with the coding and instruction tuning model. 

\begin{figure}
    \centering
    \vspace{-0.6cm}
    \begin{minipage}[b]{0.48\textwidth}
        \centering
        \includegraphics[width=\textwidth]{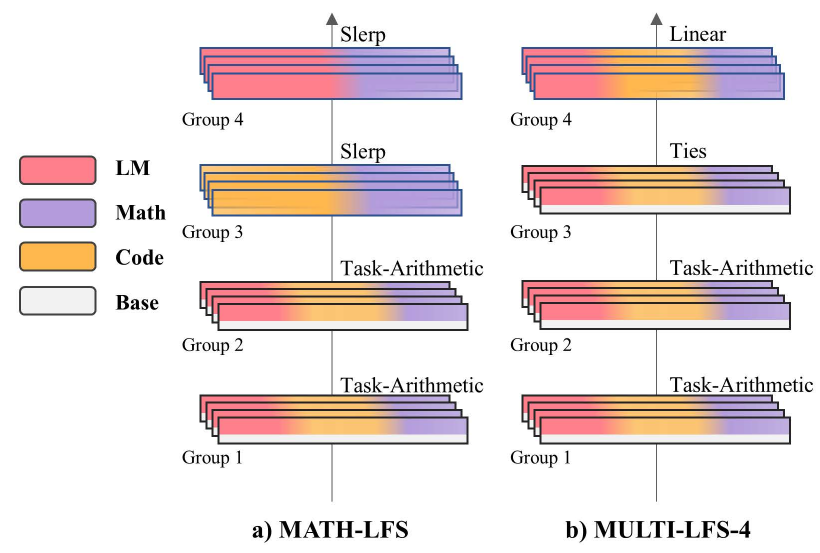}
        \captionof{subfigure}{Visualization of LFS-searched neural architectures: (a) MATH-LFS optimized for arithmetic ability; (b) MULTI-LFS-4 optimized for multiple objectives.}
        \label{fig:fig3}
    \end{minipage}
    \hfill
    \begin{minipage}[b]{0.48\textwidth}
        \centering
        \includegraphics[width=\textwidth]{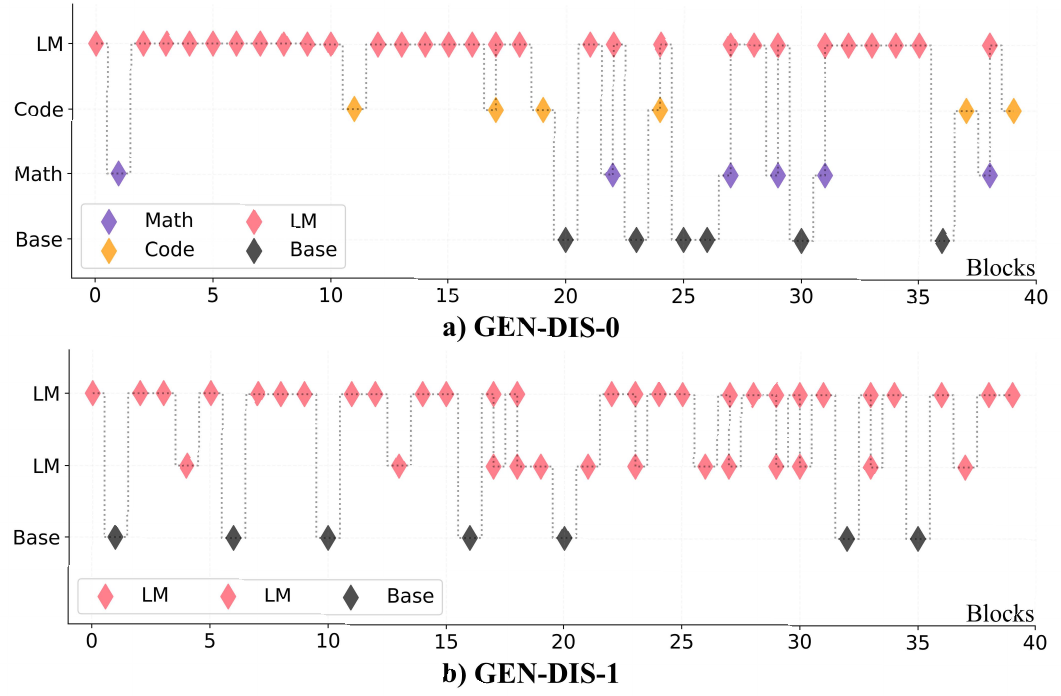}
        \captionof{subfigure}{Visualization of DIS model architecture. The x-axis represents block indices from 1 to 40, and the y-axis indicates the selected layer in each block.}
        \label{fig:fig4}
    \end{minipage}
    \label{fig:combined}
    \vspace{-0.5cm}
\end{figure}

Beyond these task-specific enhancements, we observe that LFS-searched models also demonstrate improved performance in other reasoning capabilities, with MATH-LFS showing particular strength in both common reasoning and code generation tasks, while although both GSM8K and MATH are mathematical tasks, they differ in complexity and scenarios. The merging process appears to overemphasize arithmetic, which is less relevant for MATH, leading to the observed performance drop. In comparison with alternative merging methods including TIES, Task Arithmetic, and linear merging (with hyperparameter search ranges detailed in Table~\ref{tab:table6}), our LFS method achieves superior performance on the targeted optimization objectives, demonstrating the effectiveness of our approach in single task optimization. The architecture of MATH-LFS, as illustrated in Figure  \ref{fig:fig3}, divides the model layers into four groups (G=4), where Task Arithmetic merge is employed for groups 1 and 2, while the subsequent layer groups 3 and 4 utilize SLERP merge with different model combinations - Math-Code and Math-LM respectively. The full architecture parameters can be found in Table~\ref{tab:table9}. Notably, we found that HumanEval~\citep{chen2021codex} is very sensitive to parameter changes. When merging the LM with other models, performance dropped for all methods. This may be due to the nature of the task and the small test set of only 164 samples.

\begin{table*}[t]
\centering
\scriptsize
\setlength{\tabcolsep}{0.8pt}
\caption{Performance of models merged via LFS and DIS on other Reasoning Tasks. LFS+DIS represents using LFS-merged model as an additional source model for DIS search. Numbers in blue indicate improvements over the best source model.}
\begin{tabular}{@{}l@{\hspace{0.7em}}|@{\hspace{0.7em}}c@{\hspace{0.7em}}c@{\hspace{0.7em}}c@{\hspace{0.7em}}c|@{\hspace{1.2em}}l@{\hspace{1.2em}}l@{\hspace{1.2em}}l@{}}
\toprule
\multirow{2}{*}{\textbf{Benchmark}} & \multicolumn{4}{@{\hspace{0.7em}}c|@{\hspace{1.2em}}}{\textbf{Source Models}} & \multicolumn{3}{l}{\textbf{Merging Results}} \\
\cmidrule[0.4pt]{2-5} \cmidrule[0.4pt]{6-8}
& \makebox[1cm]{\textbf{Math}} & \makebox[1cm]{\textbf{Code}} & \makebox[1cm]{\textbf{LM}} & \makebox[1cm]{\textbf{LM\_JA}} & \makebox[1.4cm]{\textbf{LFS}} & \makebox[1.4cm]{\textbf{DIS}} & \makebox[1.6cm]{\textbf{LFS+DIS}} \\
\midrule
LogiQA          & 29.65 & 25.81 & 28.57 & -- & 30.88 \textcolor{blue}{(+1.23)} & 29.34 \textcolor{gray}{(-0.31)} & \textbf{31.64} \textcolor{blue}{(+1.99)}\\
OpenBookQA      & 34.20 & \textbf{34.80} & 34.40 & -- & 37.60 \textcolor{blue}{(+2.80)}& 37.40 \textcolor{blue}{(+2.60)} & \textbf{38.60} \textcolor{blue}{(+3.80)}\\
OpenBookQA+f    & 39.20 & 44.20 & \textbf{46.20} & -- & 47.00  \textcolor{blue}{(+0.80)} & 47.40 \textcolor{blue}{(+1.20)} & \textbf{47.80} \textcolor{blue}{(+1.60)}\\
PIQA            & 79.77 & 79.92 & \textbf{79.95} & -- & \textbf{80.93} \textcolor{blue}{(+0.98)} & 79.02 \textcolor{gray}{(-0.93)} & 80.03 \textcolor{blue}{(+0.08)}\\
SocialIQA       & 46.82 & 46.77 & \textbf{51.09} & -- & \textbf{51.14} \textcolor{blue}{(+0.05)}& 50.50 \textcolor{gray}{(-0.59)} & 50.17 \textcolor{gray}{(-0.92)}\\
MGSM\_JA        & 8.00  & 4.00  & \textbf{10.40} & -- & 17.60 \textcolor{blue}{(+7.20)} & 15.20 \textcolor{blue}{(+4.80)} & \textbf{22.00} \textcolor{blue}{(+10.60)}\\
MGSM\_JA        & 8.00  & --  & -- & \textbf{8.80} & 16.40 \textcolor{blue}{(+7.60)} & \textbf{21.60} \textcolor{blue}{(+12.80)} & 19.60 \textcolor{blue}{(+10.80)}\\
\bottomrule
\multicolumn{8}{l}{\textsuperscript{*}LM\_JA denotes ELYZA-japanese-Llama-2-13b}\\
\end{tabular}
\label{tab:table1}
\vspace{-.5cm}
\end{table*}

\textbf{Depth-wise Integration} We evaluated two DIS search configurations: Configuration 1 using three candidate models (Code, Math, and LM) with depth granularity 1 and repeat factor 1; and Configuration 2 using a single candidate model (the corresponding source model for specific objective) with depth granularity 1 and repeat factor 2. Using these configurations, we similarly optimized for mathematical, coding, and general reasoning tasks respectively. DIS demonstrated effectiveness only in general reasoning tasks, yielding GEN-DIS-0 and GEN-DIS-1. Our results, as presented in Table~\ref{tab:table0}, demonstrate that both models achieved improvements,  exceeding 1\% over the source models on the MMLU benchmark. GEN-DIS-0 exhibited enhanced performance not only on the optimization objective but also demonstrated benefits on auxiliary tasks, showing improvement on the GSM8K. A detailed breakdown of performance across MMLU categories is presented in Table \ref{tab:table7}.

We visualize the network in Figure~\ref{fig:fig4}. The complete set of architectural parameters can be found in Table~\ref{tab:table10}. GEN-DIS-0 exhibited a predominantly stable selection of LM layers in early layers, gradually incorporating other source model layers with stacking behavior in middle and later layers.  Similarly, GEN-DIS-1 demonstrated no layer repetition in early layers but showed emergence of layer repetition patterns after layer $15$. Notably, although improvements were observed in common reasoning task, we found no superior configurations for mathematical and code reasoning tasks, suggesting varying sensitivity to layer ordering across different tasks.  To further explore our methodology, we extend our investigation to additional tasks.

\textbf{Expanding to other tasks} We further evaluated our method on diverse reasoning tasks including LogicQA~\citep{liu2020logiqa}, OpenBookQA~\citep{mihaylov2018can}, PIQA~\citep{bisk2020piqa}, SocialIQA~\citep{sap2019socialiqa}, and MGSM Japanese~\citep{sap2019socialiqa}, using validation datasets for searching and test sets for evaluation. For OpenBookQA, we tested with and without relevant facts in the prompt. We incorporated ELYZA-japanese-Llama-2-13b~\citep{elyzallama2023} alongside our base models (LM, Math, Code) for MGSM Japanese tasks.  See Section~\ref{sec:a6} for implementation details.  Results are presented in Table~\ref{tab:table1}. For most tasks, LFS+DFS shows promising results with significant improvements across benchmarks, notably 3.80\% on OpenBookQA and up to 10.8 points on MGSM\_JA. In addition to LFS that consistently delivers positive improvements, DFS shows occasional degradation in LogiQA and PIQA. The results particularly highlight the effectiveness of our method in cross-lingual reasoning tasks, DIS demonstrates notable effectiveness with improvements up to 12.80 points, We hypothesize DIS's preservation of processing blocks may be more effective when merging models from complementary domains. The architectural parameters can be found in Tables~\ref{tab:table11} and~\ref{tab:table12}.

\textbf{Comparison with Evolutionary-based Searching.} 
For a fair comparison, we conducted controlled experiments within our Layer-wise Fusion Search Space (LFS), evaluating our optimizer against Covariance matrix adaptation evolution strategy (CMA-ES)\citep{hansen2006cma}, as used in~\citep{akiba2024evolutionary}. We implemented CMA-ES with a 200-trial budget, matching our search budget (approximately 161 full-budget trials in MATH-LFS, see Table~\ref{tab:table2}). We optimized for mathematical, coding, and general reasoning tasks, respectively, yielding MATH-LFS-EVO, CODE-LFS-EVO, and GEN-LFS-EVO models, which we compared against our MATH-LFS, CODE-LFS, and GEN-LFS models. AS shown in Table~\ref{tab:table0}, our approach consistently outperformed the evolutionary method across all tasks, demonstrating that our multi-fidelity optimizer enables more effective search space exploration.

\subsubsection{Multi objective Optimization} Given the task-specific nature of DIS search and the more robust merging capabilities of LFS, we conducted our multi objective search on LFS with three optimization objectives: mathematical reasoning, code generation, and general reasoning. Our optimization yielded five Pareto-optimal solutions, denoted as MULTI-LFS-0 to MULTI-LFS-4, with their performance metrics presented in Table~\ref{tab:table0}. These solutions achieved significant improvements, ranging from 5.30 to 6.86 points on average compared to the best candidate model. When comparing our results with existing merging approaches such as Ties, task arithmetic, and linear merging, our method achieves better trade-offs along the Pareto frontier while maintaining higher average performance across all benchmarks. This demonstrates the inherent multi-task nature of large models and effectiveness of our multi-objective optimization strategy in finding superior model configurations that balance diverse task requirements.


As visualized in Figure~\ref{fig:fig3}, our analysis of MULTI-LFS-4 reveals an interesting layer-wise merging pattern: Task Arithmetic is optimal for groups 1 and 2, while groups 3 and 4 employ Ties and linear merging strategies. Unlike the single-objective search-derived MATH-LFS, MULTI-LFS-4 shows a preference for incorporating layers from all source models during the merging process, resulting in better preservation of comprehensive information. Experiments with fixed merging methods (only TIES or Task Arithmetic) show performance drops, with details in table \ref{tab:multi_obj_performance}.

\begin{table*}[htbp]
  \centering
  \begin{minipage}[t]{0.49\textwidth}
    \centering
    \caption{Statistics of searching trials across budget levels for MATH-LFS and GEN-DIS-1.}
    \label{tab:table2}
    \scriptsize
    \setlength{\tabcolsep}{4pt}
    \begin{tabular}{@{}l|ccc|cccc@{}}
    \toprule
    & \multicolumn{3}{c|}{\textbf{MATH-LFS}} & \multicolumn{4}{c}{\textbf{GEN-DIS-1}} \\
    & 100 & 300 & 1000 & 0 & 100 & 300 & 700 \\
    \midrule
    Trail Count & 263 & 152 & 85 & 196 & 311 & 307 & 186 \\
    Percentage (\%) & 52.6 & 30.4 & \textbf{17.0} & 19.6 & 31.1 & 30.7 & \textbf{18.6} \\
    \bottomrule
    \end{tabular}
  \end{minipage}%
  \hfill
  \begin{minipage}[t]{0.49\textwidth}
    \centering
    \caption{Ablation study on granularity ($G$) and component groups ($C$) for MATH-LFS search.}
    \label{tab:table3}
    \scriptsize
    \setlength{\tabcolsep}{6pt}
    \renewcommand{\arraystretch}{0.8} 
    \begin{tabular}{c|cccc}
       \toprule
       \multirow{2}{*}{\textbf{Component Groups}} & \multicolumn{4}{c}{\textbf{Layer Groups} ($G$)} \\
       \cmidrule{2-5}
       & 10 & 4 & 2 & 1 \\
       \midrule
       $C$=3 & 66.79 & \textbf{68.46} & 67.82 & 66.41 \\
       $C$=1 & 66.41 & 68.08 & 67.24 & 66.41 \\
       \bottomrule
   \end{tabular}
  \end{minipage}
  \vspace{-.2cm}
\end{table*}


\subsubsection{Efficiency Analysis}

Our Multi-Fidelity Optimizer dynamically adjusts budget allocation during search, with budget defined as validation dataset size. Here, we analyze the budget distribution during the search process using MATH-LFS and GEN-DIS-1 as examples (see Table \ref{tab:table8} for complete details). As shown in Table~\ref{tab:table2}, For MATH-LFS(500 trials with budgets of 100, 300, and 1000), Only 17\% of trials used the full budget, while 52\% used the minimum budget. For GEN-DIS-1, we set a maximum layer constraint of $50$ to manage computational costs, with configurations exceeding this receiving zero budget. Of 1000 initial trials, 804 were effective, with only 18.6\% using full evaluation budget. This approach significantly reduced computational costs while maintaining solution quality.


\section{Ablations and Analysis}



\textbf{Impact of granularity in LFS.}
To examine the effect of granularity in layer-wise fusion search space, we conducted ablation studies to examine the effect of granularity in LFS by varying the number of layer groups ($G$) and component groups ($C$). Table~\ref{tab:table3} shows that increasing $G$ from $1$ to $4$ consistently improves GSM8K accuracy, demonstrating the benefits of more fine-grained layer control. However, performance decreases when $G$ reaches $10$, likely due to the growth in search space exceeding our search algorithm's capability within the given trials. Our analysis also shows that increasing $C$ from $1$ to $3$ improves performance, though these gains are smaller compared to layer-wise refinement.

\begin{table*}[htbp]
  \centering
  \vspace{-0.2cm}
  \begin{minipage}[t]{0.49\textwidth}
    \centering
    \caption{Ablation study on layer retention for GEN-DIS-1 model. GEN-DIS-1-NR denotes GEN-DIS-1 without layer retention.}
    \label{tab:table4}
    \scriptsize
    \renewcommand{\arraystretch}{1.1} 
    \setlength{\tabcolsep}{1pt}
    \begin{tabular}{c|lllll}
\toprule
\textbf{Model} & \textbf{MMLU} & \textbf{GSM8K} & \textbf{MATH} & \textbf{MBPP} & \textbf{HumanEval} \\
\midrule
LM & \textbf{53.43} & 3.79 & 0.00 & 33.40 & 38.41 \\
GEN-DIS-1 & \textbf{54.76}\textcolor{blue}{↑}  & 1.74 & 0.06 & 16.80 & 18.29 \\
GEN-DIS-NR & \textbf{53.68}\textcolor{blue}{↑}  & 1.82 & 0.02 & 9.80 & 4.88 \\
\bottomrule
\end{tabular}
  \end{minipage}%
  \hfill
  \begin{minipage}[t]{0.49\textwidth}
    \centering
    \caption{Performance comparison of source models and SFS-search results}
    \label{tab:table5}
    \scriptsize
    \setlength{\tabcolsep}{1pt}
    \renewcommand{\arraystretch}{0.6} 
    \begin{tabular}{@{}l@{\hspace{5pt}}|@{\hspace{5pt}}l@{\hspace{5pt}}l@{\hspace{5pt}}l@{\hspace{5pt}}|@{\hspace{5pt}}l@{\hspace{5pt}}l@{\hspace{5pt}}l@{}}
\toprule
\textbf{Benchmark} & \multicolumn{3}{c|}{\textbf{Source Models}} & \multicolumn{3}{c}{\textbf{SFS Results}} \\
\cmidrule{2-7}
& \textbf{Math} & \textbf{Code} & \textbf{LM} & \textbf{SFS-0} & \textbf{SFS-1} & \textbf{SFS-2} \\
\midrule
MMLU & 52.04 & 52.79 & \textbf{53.43} & 52.03 & 51.91 & \textbf{53.63}\textcolor{blue}{↑}\\
GSM8K & \textbf{64.22} & 0.00 & 3.79 & \textbf{63.91}\textcolor{gray}{↓} & 0.00 & 6.52 \\
MATH & 13.70 & 0.00 & 0.00 & 13.58 & 0.00 & 0.08 \\
MBPP & 18.20 & \textbf{27.20} & 33.40 & 18.20 & \textbf{28.60}\textcolor{blue}{↑} & 30.40 \\
HumanEval & 7.32 & 23.17 & 38.41 & 7.32 & 24.39 & 34.15 \\
\bottomrule
\end{tabular}
  \end{minipage}
  \vspace{-0.2cm}
\end{table*}





\textbf{Impact of layer retention in DIS.} To explore the layer retention strategy in our DIS search space, we conducted a comparative experiment. Specifically, we modified the DIS-GEN-1 configuration by replacing the layer retention strategy with direct layer deletion, resulting in the DIS-GEN-NR.
As shown in Table \ref{tab:table4},  although it marginally outperformed the baseline language model on MMLU, its performance still fell short compared to our proposed layer retention approach. This result shows that directly deleting layers degrades model performance while retaining layers is more effective.


\textbf{Search only for rescaling.}  Several findings show that finetuning~\citep{zhang2024interpreting} or scaling~\citep{christ2024math} task-specific neurons can improve model performance. To evaluate these claims and to verify that our DIS search space (which includes scales) is not simply rescaling layers, we further evaluate Scale-Factor Search Space (SFS) that optimizes tasks by searching scaling factors to weights and layer outputs. As before, we apply our framework and obtain three SFS models: SFS-MATH, SFS-CODE, and SFS-GEN, each initialized from specialized base models. As shown in Table~\ref{tab:table5}, results show a decline in mathematical performance and slight improvements in code and reasoning tasks, though gains are modest compared to LFS and DIS, showing that scale optimization alone is not sufficient to explain the DIS effectiveness. 

\section{Conclusions and Future Work}


We presented a multi-fidelity framework for automated model merging with two complementary search spaces: Layer-wise Fusion Search for fine-grained layer merging and Depth-wise Integration Search for optimizing sequential layer arrangements. We show automated model merging not only works, but is quite effective, demonstrating strong performance in both single-objective and multi-objective scenarios, achieving a 4.24\% improvement on the GSM8K challenge task with only 17\% of the full budget within 500 trials, and a 6.86\% improvement in multi-objective performance using 18.6\% of the full budget within 1000 trials. When extended to various benchmarks, our method consistently shows promising results without any additional tuning. Overall, our work provides an efficient and flexible framework for automated model merging that achieves effective improvements with reduced computational costs. However, merged models may inherit biases from source models, requiring thorough evaluation before deployment.


\bibliographystyle{plainnat} 
\bibliography{refer}








\appendix
\section{Detailed Experimental Settings}
\subsection{Details of TIES Configuration for Math and Code Model Merging on GSM8K} \label{sec:a}
\begin{table}[htbp]
\centering
\caption{Performance Comparison with Different Parameters of Ties Merging Method}
\setlength{\tabcolsep}{4pt}
\begin{tabular}{l|cccccc}
\toprule
\textbf{Parameters} & \multicolumn{6}{c}{\textbf{Configuration}} \\
\midrule
\textbf{Ratio to retain parameters} & 0.7 & 0.5 & 0.7 & 0.5 & 0.9 & 0.9 \\
\textbf{Scaling Coefficient} & 1.0 & 0.5 & 0.5 & 1.0 & 0.5 & 1.0 \\
\midrule
\textbf{Performance on GSM8k} & 61.94 & 52.08 & 46.78 & 64.52 & 34.87  & 49.05  \\
\bottomrule
\end{tabular}
\label{tab:ties_performance}
\end{table}
\subsection{Descriptions of Existing Model Merging Methods in Layer-wise Fusion Search Space}  \label{sec:a0}
\textbf{Task Arithmetic} enhance model capabilities through vector operations by leveraging weighted combinations of task-specific knowledge. Given a base model with weights $\theta_{\text{pre}}$ and task-specific fine-tuned weights $\{\theta_{t}^{\text{ft}}\}_{t=1}^n$, task vectors are defined as:

\begin{equation}
\tau_t = \theta_{t}^{\text{ft}} - \theta_{\text{pre}}
\end{equation}

The merged weights are then computed through:

\begin{equation}
\theta_{\text{Merge}} = \theta_{\text{pre}} + \lambda \sum_{t=1}^n \tau_t
\end{equation}

where $\lambda$ controls the magnitude of task-specific adaptations.

\textbf{TIES-Merging} is a parameter conflict resolution approach that operates in three stages. First, select the top $k\%$ parameters by magnitude of each task vector $\tau_t$:

\begin{equation}
\hat{\tau}_t = \text{TopK}(\tau_t, k)
\end{equation}

Next, Generating a consensus sign vector by examining the aggregate direction of parameter changes across all tasks:

\begin{equation}
\hat{\gamma} = \text{sgn}\left(\sum_{t=1}^n \hat{\tau}_t\right)
\end{equation}

Finally, computing the average update magnitude considering only those task vectors whose signs align with the consensus direction:

\begin{equation}
\tilde{\tau} = \text{Average}(\{\hat{\tau}_t : \text{sgn}(\hat{\tau}_t) = \hat{\gamma}\})
\end{equation}

The final merged model weights are then computed as:

\begin{equation}
\theta_{\text{Merge}} = \theta_{\text{pre}} + \lambda * \tilde{\tau}
\end{equation}

\textbf{SLERP} (Spherical Linear Interpolation ) computes optimal geodesic paths between model weights through:

\begin{equation}
\text{SLERP}(\theta_1, \theta_2, t) = \frac{\sin((1-t)\omega)}{\sin(\omega)}\theta_1 + \frac{\sin(t\omega)}{\sin(\omega)}\theta_2
\end{equation}

where $\omega = \arccos\left(\frac{\langle\theta_1, \theta_2\rangle}{\|\theta_1\|\|\theta_2\|}\right)$ and $t \in [0,1]$ is the interpolation parameter.

\textbf{Linear Merging} implements straightforward weighted averaging:

\begin{equation}
\theta_{\text{Linear}} = \sum_{t=1}^n w_t\theta_t
\end{equation}

where $\sum_{t=1}^n w_t = 1$ and $w_t \geq 0$.


\subsection{Descriptions of SMAC-based Multi-Fidelity Optimization} \label{sec:a1}
Our implementation extends SMAC~\citep{JMLR:v23:21-0888}, integrating Hyperband (HB)~\citep{li2018hyperband} with Bayesian Optimization (BO)~\citep{snoek2012practical} and employing Random Forest~\citep{breiman2001random} as the surrogate model.

The framework operates using minimum and maximum budgets ($b_{\text{min}}$, $b_{\text{max}}$) with a spacing parameter $\eta > 1$. The algorithm creates $s_{\text{max}} = \lfloor\log_\eta(b_{\text{max}}/b_{\text{min}})\rfloor$ brackets, each initiating with $n_i = \lfloor\eta^{s_{\text{max}}-i} \cdot \frac{\eta}{\eta-1}\rfloor$ configurations. Within each bracket, Successive Halving proceeds through $\lfloor \log_\eta(\frac{n_i}{n_{\text{min}}}) \rfloor + 1$ rounds, evaluating configurations at increasing budgets while progressively eliminating underperforming candidates. Specifically, after evaluating all configurations at budget $b$, only the top $\lfloor \frac{n_i}{\eta^l} \rfloor$ performers advance to the next round with an increased budget of $\eta b$.

A key enhancement is the Random Forest model that learns from all prior configuration-performance pairs, prioritizing data from higher budgets. This model guides the selection of promising configurations via Expected Improvement, balancing exploration and exploitation. As the optimization progresses, the evaluation of more configurations at higher budgets enables the algorithm to correct potential misjudgments from lower-fidelity evaluations.

\looseness -1 For a detailed algorithmic description, see Algorithm~\ref{alg:smac}, which presents the complete optimization process incorporating trial limits. This integration of multi-fidelity resource allocation with surrogate-based modeling delivers efficient configuration space exploration while maintaining evaluation quality.

\begin{algorithm}
\caption{SMAC-based Multi-Fidelity Optimization}
\label{alg:smac}
\begin{algorithmic}[1]
\Require Configuration space $\Theta$, minimum budget $b_{\text{min}}$, maximum budget $b_{\text{max}}$, spacing factor $\eta > 1$, maximum trials $T_{\text{max}}$
\Ensure Optimized configuration $\theta^*$

\State $s_{\text{max}} \gets \lfloor\log_\eta(\frac{b_{\text{max}}}{b_{\text{min}}})\rfloor$ \Comment{Maximum brackets}
\State $\mathcal{D} \gets \emptyset$ \Comment{Observation history}
\State $\theta^* \gets \emptyset$, $y^* \gets \infty$ \Comment{Best configuration tracking}
\State $T \gets 0$ \Comment{Initialize trial counter}

\For{$i \in \{s_{\text{max}}, s_{\text{max}}-1, \ldots, 0\}$}
    \If{$T \geq T_{\text{max}}$}
        \State \textbf{break} \Comment{Exit if reached maximum trials}
    \EndIf
    
    \State $n_i \gets \lfloor\eta^{s_{\text{max}}-i} \cdot \frac{\eta}{\eta-1}\rfloor$ \Comment{Initial configurations}
    \State $\mathcal{M} \gets \text{FitRandomForest}(\mathcal{D})$ \Comment{Build surrogate model}
    
    \If{$|\mathcal{D}| = 0$}
        \State $\Theta_i \gets$ Sample $n_i$ random configurations from $\Theta$
    \Else
        \State $\Theta_i \gets$ Select $n_i$ configurations with highest EI based on $\mathcal{M}$
    \EndIf
    
    \State $s_i \gets \lfloor \log_\eta(\frac{n_i}{1}) \rfloor + 1$ \Comment{SH rounds}
    \State $\mathcal{A} \gets \Theta_i$ \Comment{Set of active configurations}
    \State $b \gets b_{\text{min}} \cdot \eta^i$ \Comment{Initial budget}
    
    \For{$l \in \{0, 1, \ldots, s_i-1\}$}
        \If{$T \geq T_{\text{max}}$}
            \State \textbf{break} \Comment{Exit if reached maximum trials}
        \EndIf
        
        \State $n_{i,l} \gets \lfloor \frac{n_i}{\eta^l} \rfloor$ \Comment{Current pool size}
        
        \For{each $\theta \in \mathcal{A}$}
            \State $y_{\theta} \gets f(\theta, b)$ \Comment{Evaluate configuration}
            \State $\mathcal{D} \gets \mathcal{D} \cup \{(\theta, b, y_{\theta})\}$ \Comment{Update history}
            \State $T \gets T + 1$ \Comment{Increment trial counter}
            
            \If{$b = b_{\text{max}}$ and $y_{\theta} < y^*$}
                \State $y^* \gets y_{\theta}$, $\theta^* \gets \theta$ \Comment{Update best}
            \EndIf
            
            \If{$T \geq T_{\text{max}}$}
                \State \textbf{break} \Comment{Exit if reached maximum trials}
            \EndIf
        \EndFor
        
        \State Sort $\mathcal{A}$ by performance
        \State $\mathcal{A} \gets$ Top $\lfloor \frac{n_{i,l}}{\eta} \rfloor$ configurations from $\mathcal{A}$
        \State $b \gets \min(b \cdot \eta, b_{\text{max}})$ \Comment{Increase budget}
        
        \If{$b = b_{\text{max}}$ or $|\mathcal{A}| = 1$}
            \State \textbf{break}
        \EndIf
    \EndFor
\EndFor

\State \Return $\theta^*$
\end{algorithmic}
\end{algorithm}

\subsection{Details of Datasets Information} \label{sec:a2}
For searching, We use data from three reasoning domains: $1,000$ GSMPlus~\citep{li2024gsm} samples, an adversarial variant of GSM8K with mathematical perturbations for testing math reasoning; $300$ MBPP~\citep{austin2021program} samples ($200$ training/$100$ validation) for code understanding; and $700$ samples from MMLU validation\citep{hendrycks2020measuring} for general reasoning. These datasets support both single-objective optimization when used separately and multi-objective optimization when combined. For comprehensive performance evaluation, we employ established benchmark test sets across three key domains: mathematical reasoning using the complete test sets from GSM8K~\citep{cobbe2021training} and MATH~\citep{hendrycksmath2021}, code generation using the standard test splits from MBPP~\citep{austin2021program} and HumanEval~\citep{chen2021codex}, and general knowledge using the  MMLU test set~\citep{hendrycks2020measuring}, which spans diverse knowledge domains.

\subsection{Evaluation Metrics and details} \label{sec:a3}
We evaluate using domain-specific metrics: accuracy for MMLU multiple choice questions based on loglikelihood, zero-shot accuracy for GSM8K and MATH, and Pass@1 for HumanEval and MBPP. We run all evaluations using LM Evaluation Harness\citep{eval-harness} with vLLM\citep{kwon2023efficient} acceleration. For consistency, we use fixed parameters across all tests: batch size $16$, temperature $0.0$ for greedy decoding, and maximum generation length of $1,024$ tokens for GSM8K and $2,048$ tokens for other datasets. All experiments run on NVIDIA Tesla A100 GPUs.

\subsection{Details of Search Spaces} \label{sec:a4}
Within LFS, we define specific parameter ranges for each merging method: Task Arithmetic utilizes task vector weights $\lambda \in [0,1]$; TIES-Merging combines task vector weights $\lambda \in [0,1]$ with a ratio to retain parameters $k \in [0.1,0.99]$; Linear-merging optimizes model coefficients $w_t \in [0,1]$ subject to $\sum_i w_t = 1$; and Slerp employs an interpolation parameter $t \in [0,1]$.

\subsection{Details of Grid Search on Hyperparameters of base Model Merging Methods} \label{sec:a5}

\begin{table}[htbp]
\centering
\renewcommand{\arraystretch}{0.7}  
\caption{Search Ranges of Hyperparameters for Different Model Merging Methods}
\begin{tabular}{l|p{10cm}}
\toprule
\textbf{Merging Methods} & \textbf{Search Ranges of Hyperparameters} \\
\midrule
Task Arithmetic & Scaling term to merge parameters: [0.5, 1.0] \\
\midrule
Linear Merging & Scaling term to merge parameters: $[1/n]$ ($n$: number of models) \\
\midrule
TIES-Merging & \begin{tabular}[t]{@{}l@{}}
Scaling term to merge parameters: [0.5, 1.0]\\[0.5ex]
Ratio to retain parameters with largest-magnitude values: [0.5, 0.7, 0.9]
\end{tabular} \\
\bottomrule
\end{tabular}
\label{tab:table6}
\end{table}

\subsection{Details of search on other Reasoning Tasks} \label{sec:a6}
\textbf{LogiQA} is derived from the logical comprehension section of China's National Civil Servants Examination, specifically designed to evaluate candidates' critical thinking and problem-solving capabilities. For our search implementation, we utilize the validation dataset with a budget range of $100-651$.

\textbf{OpenBookQA} is used to measure deep understanding of both subject matter and language comprehension. The dataset comes with an "open book" of fundamental facts. We conducted experiments both with and without facts in the prompt. Our search employs the validation dataset with a budget range of $100-500$.

\textbf{PIQA (Physical Interaction: Question Answering)} serves as a benchmark dataset for physical commonsense reasoning, with a particular focus on everyday situations and unconventional solutions. We have sampled 1,000 examples from the validation dataset for our search purposes, setting the budget range at $100-1,000$.

\textbf{SocialIQA} stands as a comprehensive benchmark for testing social commonsense intelligence, this dataset evaluates understanding of human actions and their social implications in everyday situations. Our search implementation uses a 1,000-sample subset from the validation data, with a budget range of $100-1,000$.

\textbf{MGSM (Multilingual Grade School Math Benchmark)} is a benchmark of grade-school math problems The same 250 problems from GSM8K are each translated via human annotators in 10 languages. 
we use 1,069 mathematics problems and solutions translated to japanese from the GSM8K test set by Sakana AI for searching ,set a budget range of $100-1000$.

\section{Additional Experimental Results.}
\subsection{Detailed breakdown of performance across specific MMLU categories of GEN-DIS-0 and GEN-DIS-1} \label{sec:b0}

\begin{table}[H]
\centering
\renewcommand{\arraystretch}{0.95}  
\caption{Performance Comparison on MMLU Subject Categories between LM and GEN-DIS}
\setlength{\tabcolsep}{4pt}
\begin{tabular}{l|ccc}
\toprule
\textbf{MMLU Category} & \textbf{LM} & \textbf{GEN-DIS-0} & \textbf{GEN-DIS-1} \\
\midrule
Social Sciences & 62.24 & 63.24 \small\textcolor{blue}{(+1.00)} & \textbf{63.69} \small\textcolor{blue}{(+1.45)} \\
Humanities & 49.52 & \textbf{51.31} \small\textcolor{blue}{(+1.79)} & 51.09 \small\textcolor{blue}{(+1.57)} \\
STEM & 42.82 & 43.67 \small\textcolor{blue}{(+0.85)} & \textbf{44.37} \small\textcolor{blue}{(+1.55)} \\
Other & 61.31 & \textbf{62.66} \small\textcolor{blue}{(+1.35)} & 62.12 \small\textcolor{blue}{(+0.81)} \\
\bottomrule
\end{tabular}
\label{tab:table7}
\end{table}

\subsection{Performance comparison of different merging methods in multi-objective optimization.}

\begin{table}[htbp]
\centering
\small
\caption{Performance comparison of different merging methods in multi-objective optimization}
\renewcommand{\arraystretch}{1.3} 
\setlength{\tabcolsep}{2pt}
\begin{tabular}{llcccccccc}
\hline
\textbf{Method} & \textbf{Model} & \textbf{Src$^\dagger$} & \textbf{Srch$^\ddagger$} & \textbf{MMLU} & \textbf{GSM8K} & \textbf{MATH} & \textbf{MBPP} & \textbf{HEval} & \textbf{Avg} \\
\hline
Multi-Obj & MULTI-LFS-0 & M+C+L & 1+2+3 & 55.03 & 63.08 & 11.76 & 32.60 & 21.95 & 36.88 (+5.78) \\
(Our) & MULTI-LFS-1 & M+C+L & 1+2+3 & 54.70 & 66.94 & 11.38 & 30.60 & 23.78 & 37.48 (+6.38) \\
 & MULTI-LFS-2 & M+C+L & 1+2+3 & 54.67 & 65.13 & 11.06 & 30.40 & 20.73 & 36.40 (+5.30) \\
 & MULTI-LFS-3 & M+C+L & 1+2+3 & 54.99 & 65.50 & 9.42 & 30.20 & 23.17 & 36.66 (+5.56) \\
 & MULTI-LFS-4 & M+C+L & 1+2+3 & 54.77 & 66.79 & 12.06 & 31.80 & 24.40 & 37.96 (+6.86) \\
\hline
Multi-Obj & MULTI-LFS-0-TA & M+C+L & 1+2+3 & 54.92 & 62.85 & 10.62 & 28.60 & 18.29 & 35.06 (+3.96) \\
(Task Arith) & MULTI-LFS-1-TA & M+C+L & 1+2+3 & 54.97 & 63.07 & 10.70 & 31.40 & 20.12 & 36.06 (+4.96) \\
 & MULTI-LFS-2-TA & M+C+L & 1+2+3 & 55.46 & 61.56 & 10.52 & 29.80 & 24.39 & 36.35 (+5.25) \\
 & MULTI-LFS-3-TA & M+C+L & 1+2+3 & 55.47 & 63.56 & 11.36 & 30.00 & 20.73 & 36.26 (+5.16) \\
 & MULTI-LFS-4-TA & M+C+L & 1+2+3 & 54.96 & 61.48 & 10.20 & 29.20 & 18.29 & 34.83 (+3.73) \\
\hline
Multi-Obj & MULTI-LFS-0-Ties & M+C+L & 1+2+3 & 53.51 & 63.23 & 11.64 & 29.60 & 20.14 & 36.62 (+5.52) \\
(Ties) & MULTI-LFS-1-Ties & M+C+L & 1+2+3 & 54.62 & 61.18 & 10.52 & 32.00 & 18.90 & 35.44 (+4.34) \\
 & MULTI-LFS-2-Ties & M+C+L & 1+2+3 & 54.99 & 61.56 & 11.36 & 30.00 & 20.73 & 35.73 (+4.63) \\
 & MULTI-LFS-3-Ties & M+C+L & 1+2+3 & 55.30 & 62.26 & 9.86 & 29.80 & 18.29 & 35.10 (+4.00) \\
 & MULTI-LFS-4-Ties & M+C+L & 1+2+3 & 54.24 & 61.25 & 10.70 & 30.20 & 18.85 & 35.05 (+3.95) \\
\hline
\end{tabular}

\vspace{0.5em}
\begin{flushleft}
\footnotesize{$^\dagger$Src = Source, M+C+L = Math+Code+LM}\\
\footnotesize{$^\ddagger$Srch = Search, Dataset Index: 1=GSMPlus, 2=MBPPval, 3=MMLUval}\\
\footnotesize{HEval = HumanEval, Avg = Average}
\end{flushleft}
\label{tab:multi_obj_performance}
\end{table}

\subsection{Additional result of Budget Distribution } \label{sec:b1}

\begin{table}[H]
\centering
\renewcommand{\arraystretch}{0.9} 
\setlength{\tabcolsep}{2pt}
\caption{Budget Distribution Across Models}
\renewcommand{\arraystretch}{1.2}
\begin{tabular}{l|c|c|c|c|c|c}
\toprule
\textbf{Budget} & \textbf{MATH-LFS} & \textbf{CODE-LFS} & \textbf{Multi-LFS} & \textbf{GEN-LFS} & \textbf{GEN-DIS-0} & \textbf{GEN-DIS-1} \\
\midrule
100 & 263 (47.2\%) & 295 (59.0\%) & 240 (48.0\%) & 207 (41.4\%) & 163 (16.3\%) & 311 (31.1\%) \\
300 & 152 (30.4\%) & 205 (41.0\%) & 161 (32.2\%) & 181 (36.2\%) & 165 (16.5\%) & 307 (30.7\%) \\
1000 & 85 (17.0\%) & - & 99 (19.8\%) & 112 (22.4\%) & 109 (10.9\%) & 186 (18.6\%) \\
0 & - & - & - & - & 563 (56.3\%) & 196 (19.6\%) \\
\bottomrule
\end{tabular}
\label{tab:table8}
\end{table}

\section{Comparison with evolutionary-based merging} \label{sec:compare}

\subsection{Similarity}
Both studies are hyperparameter optimization-based methods, incorporating layerwise and depthwise modifications---a common strategy in the merging community to maintain architecture stability without post-finetuning \cite{ilharco2022editing, yadav2024ties, yu2024language}.

\subsection{Difference}
However, our method is more effective, which differs in the following three key aspects:

\textbf{Search Space Design} For layerwise search, \cite{akiba2024evolutionary} only supports TIES-Merging with uniform layer merging. Ours offers flexibility through multiple merging methods, layer selection, and granular controls. This fine-grained control mitigates layer conflicts and improves performance, as shown in our ablation studies (Table \ref{tab:table3}). Our experiments reveal that different objectives exhibit varying preferences for merging methods and layer selection.
    
For depthwise search, \cite{akiba2024evolutionary} explores all possible layer interactions, resulting in an excessively large search space that makes it difficult to optimize. Our block-wise design controls the degree of layer interaction through depth granularity, which enables flexible adjustment for different tasks while reducing search space size with fewer hyperparameters (see the Table\ref{tab:search_space_comparison}). While our permutation vector $P$ grows exponentially with increasing depth granularity, we have the alternative to replace this by optimizing a learned group of parameters (with the size equal to $M \times D \times R$) and determine the order of layers by sorting its values (the complexity shifts from exponential to linear).

\begin{table}[htbp]
\centering
\caption{Comparison of search space parameters between Evolutionary Search and our approach}
\renewcommand{\arraystretch}{1.5}
\begin{tabular}{lcl}
\hline
\rowcolor[gray]{0.9}
\textbf{Definition} & \textbf{Evolutionary Search Space} & \textbf{Our Approach} \\
\hline
$L, M$ & \multicolumn{2}{l}{Number of layers/candidate models} \\
\hline
$R$ & Repetition of candidate models & Block-wise layer repetition \\
\hline
$D$ & -- & Depth granularity \\
\hline
Layer selection & $2^{L \cdot M \cdot R}$ & $2^{M \cdot D \cdot R} \cdot L/D$ \\
\hline
Scaling matrix $W$ & $(L \cdot M \cdot R)^2$ & $L/D$ \\
\hline
\multirow{2}{*}{Permutation vector $P$} & \multirow{2}{*}{--} & $L/D \cdot (D \cdot M \cdot R)!/(R!)^{M \cdot D}$ \\
& & or $M \cdot D \cdot R \cdot L/D$ \\
\hline
\end{tabular}
\label{tab:search_space_comparison}
\end{table}

\textbf{Objective}
Compared to \cite{akiba2024evolutionary} which only supports single-objective optimization, our study is extended for multi-objective optimization. This is particularly important for large models, as we observe that optimizing for a single objective often leads to improvements in other tasks, demonstrating the inherent multi-task nature of large models. Furthermore, with our multi-objective optimization, our merged models maintain excellent balance across all tasks.

\textbf{Optimizers}
\cite{akiba2024evolutionary} employs a full-budget, evolution-based optimization, which can be computationally intensive in large search spaces, while our method leverages multi-fidelity evaluation, enabling more effective exploration of the search space. This allows for more trials within the same computational budget while facilitating extensions to more complex multi-objective optimization settings.



\begin{sidewaystable}
\centering
\caption{Configuration parameters of architectures Searched by LFS}
\label{tab:lfsmodels_hyper}
\resizebox{\textwidth}{!}{
\begin{tabular}{ll|l|cccccccc}
\toprule
\textbf{Group} & \textbf{Method} & \textbf{Metric} & \textbf{MATH-LFS} & \textbf{GEN-LFS} & \textbf{MULTI-LFS-0} & \textbf{MULTI-LFS-1} & \textbf{MULTI-LFS-2} & \textbf{MULTI-LFS-3} & \textbf{MULTI-LFS-4} \\
\midrule
\multirow{6}{*}{Group (1-10)} 
& \multirow{3}{*}{task\_arithmetic} & mlp & 0.983 & -- & -- & 0.303 & 0.303 & 0.303 & 0.293 \\
& & att & 0.182 & -- & -- & 0.948 & 0.948 & 0.948 & 0.881 \\
& & other & 0.791 & -- & -- & 0.997 & 0.997 & 0.997 & 0.997 \\
\cmidrule{2-10}
& \multirow{3}{*}{linear} & mlp & -- & [0.118, 0.324, 0.558] & [0.351, 0.344, 0.304] & -- & -- & -- & -- \\
& & att & -- & [0.430, 0.224, 0.346] & [0.162, 0.298, 0.540] & -- & -- & -- & -- \\
& & other & -- & [0.317, 0.257, 0.426] & [0.288, 0.371, 0.340] & -- & -- & -- & -- \\
\midrule
\multirow{6}{*}{Group (11-20)} 
& \multirow{3}{*}{task\_arithmetic} & mlp & 0.982 & -- & 0.395 & 0.395 & 0.395 & 0.395 & 0.395 \\
& & att & 0.604 & -- & 0.842 & 0.842 & 0.842 & 0.842 & 0.862 \\
& & other & 0.329 & -- & 0.300 & 0.380 & 0.300 & 0.300 & 0.300 \\
\cmidrule{2-10}
& \multirow{3}{*}{linear} & mlp & -- & [0.376, 0.258, 0.366] & -- & -- & -- & -- & -- \\
& & att & -- & [0.414, 0.357, 0.229] & -- & -- & -- & -- & -- \\
& & other & -- & [0.263, 0.532, 0.205] & -- & -- & -- & -- & -- \\
\midrule
\multirow{9}{*}{Group (21-30)} 
& \multirow{3}{*}{slerp(Math, Code)} & mlp & 0.594 & -- & -- & -- & -- & -- & -- \\
& & att & 0.566 & -- & -- & -- & -- & -- & -- \\
& & other & 0.323 & -- & -- & -- & -- & -- & -- \\
\cmidrule{2-10}
& \multirow{3}{*}{ties} & mlp & -- & -- & 0.778/0.583 & 0.778/0.583 & 0.819/0.583 & -- & 0.724/0.583 \\
& & att & -- & -- & 0.394/0.312 & 0.394/0.312 & 0.373/0.312 & -- & 0.488/0.312 \\
& & other & -- & -- & 0.324/0.606 & 0.349/0.606 & 0.324/0.606 & -- & 0.371/0.606 \\
\cmidrule{2-10}
& \multirow{3}{*}{linear} & mlp & -- & -- & -- & -- & -- & [0.4739, 0.4728, 0.0533] & -- \\
& & att & -- & -- & -- & -- & -- & [0.3916, 0.3649, 0.2435] & -- \\
& & other & -- & -- & -- & -- & -- & [0.0342, 0.2826, 0.6832] & -- \\
\midrule
\multirow{6}{*}{Group (31-40)} 
& \multirow{3}{*}{slerp(LM, Math)} & mlp & 0.469 & -- & -- & -- & 0.113 & -- & -- \\
& & att & 0.437 & -- & -- & -- & 0.339 & -- & -- \\
& & other & 0.549 & -- & -- & -- & 0.58 & -- & -- \\
\cmidrule{2-10}
& \multirow{3}{*}{linear} & mlp & -- & -- & [0.354, 0.382, 0.264] & [0.375, 0.392, 0.234] & -- & [0.373, 0.390, 0.237] & [0.354, 0.382, 0.264] \\
& & att & -- & -- & [0.568, 0.141, 0.292] & [0.556, 0.168, 0.276] & -- & [0.562, 0.159, 0.279] & [0.548, 0.156, 0.296] \\
& & other & -- & -- & [0.344, 0.183, 0.473] & [0.345, 0.183, 0.473] & -- & [0.345, 0.183, 0.473] & [0.345, 0.183, 0.473] \\
\bottomrule
\end{tabular}
}
\label{tab:table9}
\end{sidewaystable}

\begin{table}[htbp]
\centering
\caption{DIS-Optimized Architecture Parameters for General Reasoning}
\begin{tabular}{c||c|c|c|c||c|c|c||c|c|c}
\hline
\multirow{2}{*}{block} & \multicolumn{4}{c||}{GEN-DIS-0} & \multicolumn{3}{c||}{GEN-DIS-1} & \multicolumn{3}{c}{GEN-DIS-NR} \\
\cline{2-11}
& layer\_0 & layer\_1 & layer\_2 & scale & layer\_0 & layer\_1 & scale & layer\_0 & layer\_1 & scale\\
\hline
1 &LM & -- & -- & 1.00& LM& -- &  0.99 & LM& -- & 1.07\\
2 & Math& -- & -- &1.00 & Base & -- & 0.78& LM& -- &1.00\\
3 & LM& -- & -- &1.00 & LM & -- & 0.93 & LM& -- &1.00\\
4 & LM& -- & -- &1.00 & LM& -- & 1.01 & LM& -- &1.00\\
5 & LM& -- & -- &1.00 & LM& -- & 1.00& LM& -- &1.00\\
6 &LM & -- & -- &1.00 & LM& -- & 0.93& LM& -- &1.00\\
7 & LM& -- & -- &1.00 & Base& -- & 1.10 & LM & -- &1.00\\
8 & LM& -- & -- &1.00 & LM& -- &1.00 & LM& -- &1.00\\
9 & LM& -- & -- &1.00 & LM& -- &1.14 & LM & -- &1.00\\
10 &LM & -- & -- & 1.01& LM& -- & 1.00& LM & -- &1.00\\
11 & LM& -- & -- &1.00 & Base& -- &1.00 & LM& -- & 0.97\\
12 & Code& -- & -- &1.00 & LM & -- &0.95 & LM& -- &1.00\\
13 & LM& -- & -- &1.00 & LM& -- & 0.99& LM& -- & 1.12\\
14 & LM& -- & -- &1.00& LM& -- & 1.00& LM& -- &1.00\\
15 & LM& -- & -- &1.00 & LM& -- & 0.92 & LM& -- & 1.14\\
16 & LM& -- & -- & 1.07& LM& -- & 0.89 & LM& -- & 0.89\\
17 & LM& -- & -- & 1.00& Base& -- & 1.00& LM& -- &1.00\\
18 & Code&LM & -- & 1.00& LM&LM & 1.00& LM & -- &1.00\\
19 & LM& -- & -- & 1.00& LM&LM &1.00 & LM& -- &1.00\\
20 & Code& -- & - &1.00 & LM& -- & 1.00& LM& -- &1.00\\
21 & Base& -- & -- &1.00 & Base& -- &1.00 & -- & -- & 1.15\\
22 & LM& -- & -- & 1.00&  LM& -- &1.00 & LM& -- &1.00\\
23 & Math& LM& -- & 1.00& LM& -- &1.00 & LM & -- & 1.04 \\
24 & Base& -- & -- &1.00 &LM &LM & 1.00& LM& -- &1.00\\
25 & Code& LM& -- &1.00 & LM& -- &1.00 & -- & -- & 1.00\\
26 & Base& -- & -- & 1.00& LM& -- &1.00 & -- & -- & 0.93\\
27 & Base& -- & -- & 1.05 & LM& -- & 1.05& LM& -- & 1.00\\
28 & Math& LM& -- &1.00 & LM& LM& 1.00& LM & LM& 1.06\\
29 & LM& -- & -- & 1.07 & LM& -- & 1.00& LM& -- &1.00\\
30 & Math& LM&  -- &0.87 &LM &LM &1.00 & LM& -- & 1.21\\
31 & Base & -- & -- &1.00 & LM&LM &1.00 &LM & LM&1.00\\
32 & Math&LM & -- & 1.00& LM& &1.05 & LM & -- & 1.10\\
33 & LM& -- & -- & 1.00& Base & -- &1.00 & LM & LM & 0.99\\
34 & LM& -- & -- & 1.00& LM & LM &1.00 & LM & -- & 1.00\\
35 & LM& -- & -- &1.00 & LM & -- &1.00 & LM & -- &1.08\\
36 & LM& -- & -- &1.20 & Base & -- & 1.00& LM & -- &1.07\\
37 & Base& -- & -- &1.00 & LM & -- & 1.13& LM & LM &1.32\\
38 &Code & -- & -- &1.00 & LM & -- &1.03 & LM& LM &1.00\\
39 &Math &LM & -- &1.00 & LM & -- &1.00 &LM & -- &1.00\\
40 &Code & -- & -- & 1.00& LM& --  &1.01 & LM& -- &1.00\\
\hline
\end{tabular}
\label{tab:table10}
\end{table}

\begin{table}[htbp]
\centering
\caption{DIS-Optimized Architecture Parameters for OpenbookQA}
\begin{tabular}{c||c|c|c|c||c|c|c|c}
\hline
\multirow{2}{*}{block} & \multicolumn{4}{c||}{OpenbookQA} & \multicolumn{4}{c}{OpenbookQA+F} \\
\cline{2-9}
& layer\_0 & layer\_1 & layer\_2 & scale & layer\_0 & layer\_1 & layer\_2 & scale \\
\hline
1 & LM & -- & -- & 1.00 & Base & -- & -- & 0.96 \\
2 & Code & -- & -- & 1.00 & Code & LM & -- & 1.00 \\
3 & LM & -- & -- & 1.00 & LM & -- & -- & 1.00 \\
4 & LM & -- & -- & 1.00 & LM & -- & -- & 0.97 \\
5 & LM & -- & -- & 1.06 & LM & Code & -- & 1.00 \\
6 & LM & -- & -- & 1.00 & Base & -- & -- & 1.00 \\
7 & LM & -- & -- & 1.00 & Base & -- & -- & 1.00 \\
8 & Math & LM & -- & 1.00 & LM & -- & -- & 1.19 \\
9 & LM & -- & -- & 1.00 & Code & -- & -- & 1.00 \\
10 & LM & -- & -- & 1.00 & LM & -- & -- & 1.00 \\
11 & LM & -- & -- & 1.00 & Math & -- & -- & 1.00 \\
12 & Base & -- & -- & 1.00 & Base & -- & -- & 1.00 \\
13 & LM & -- & -- & 1.00 & LM & -- & -- & 0.91 \\
14 & LM & -- & -- & 1.00 & Base & -- & -- & 1.00 \\
15 & LM & -- & -- & 1.00 & LM & -- & -- & 1.08 \\
16 & LM & -- & -- & 1.00 & LM & -- & -- & 1.00 \\
17 & LM & -- & -- & 1.00 & LM & -- & -- & 0.99 \\
18 & LM & -- & -- & 0.85 & LM & -- & -- & 1.00 \\
19 & LM & -- & -- & 1.00 & LM & -- & -- & 1.00 \\
20 & LM & -- & -- & 1.00 & Code & -- & -- & 1.07 \\
21 & LM & -- & -- & 1.00 & Base & -- & -- & 1.21 \\
22 & LM & -- & -- & 1.00 & LM & -- & -- & 0.98 \\
23 & LM & -- & -- & 1.00 & Code & -- & -- & 1.00 \\
24 & LM & -- & -- & 1.00 & Base & -- & -- & 0.92 \\
25 & Code & LM & -- & 1.00 & LM & -- & -- & 1.00 \\
26 & Base & -- & -- & 1.00 & LM & -- & -- & 0.93 \\
27 & Code & LM & -- & 1.00 & Math & LM & -- & 1.08 \\
28 & Code & LM & -- & 0.96 & Code & -- & -- & 1.18 \\
29 & LM & -- & -- & 1.00 & LM & -- & -- & 1.00 \\
30 & Math & LM & -- & 1.00 & Code & LM & -- & 1.01 \\
31 & LM & -- & -- & 1.00 & LM & -- & -- & 0.89 \\
32 & LM & -- & -- & 1.00 & Base & -- & -- & 1.05 \\
33 & LM & -- & -- & 1.00 & LM & -- & -- & 1.00 \\
34 & LM & -- & -- & 1.00 & Math & LM & -- & 0.83 \\
35 & LM & -- & -- & 1.00 & Code & LM & -- & 1.12 \\
36 & LM & -- & -- & 1.20 & LM & -- & -- & 1.00 \\
37 & LM & -- & -- & 1.00 & Math & -- & -- & 1.13 \\
38 & LM & -- & -- & 1.00 & LM & -- & -- & 1.03 \\
39 & LM & -- & -- & 1.00 & Math & -- & -- & 1.00 \\
40 & LM & -- & -- & 1.00 & Base & -- & -- & 1.01 \\
\hline
\end{tabular}
\label{tab:table11}
\end{table}

\begin{table}[htbp]
\centering
\caption{DIS-Optimized Architecture Parameters for MGSM\_JA}
\begin{tabular}{c||c|c|c|c||c|c|c}
\hline
\multirow{2}{*}{block} & \multicolumn{4}{c||}{ MGSM\_JA\_0} & \multicolumn{3}{c}{MGSM\_JA\_1} \\
\cline{2-8}
& layer\_0 & layer\_1 & layer\_2 & scale & layer\_0 & layer\_1 & scale \\
\hline
1 & Base & -- & -- & 0.99 & Math & -- & 0.98 \\
2 & Math & -- & -- & 1.00 & Base & -- & 1.12 \\
3 & Math & -- & -- & 0.96 & LM\_JA & -- & 1.00 \\
4 & Math & -- & -- & 1.00 & Base & -- & 1.00 \\
5 & Math & -- & -- & 1.00 & Math & LM\_JA & 1.00 \\
6 & Math & -- & -- & 1.00 & Math & -- & 0.93 \\
7 & Math & LM & -- & 1.00 & Math & -- & 1.00 \\
8 & Math & -- & -- & 1.11 & Math & -- & 1.00 \\
9 & Math & LM & -- & 1.00 & Math & -- & 1.00 \\
10 & Math & -- & -- & 1.10 & Math & LM\_JA & 1.00 \\
11 & Math & -- & -- & 1.00 & LM\_JA & -- & 1.00 \\
12 & Math & LM & -- & 1.06 & Math & LM\_JA & 1.00 \\
13 & LM & Math & -- & 1.00 & Math & -- & 1.00 \\
14 & LM & Math & -- & 0.83 & Math & -- & 0.97 \\
15 & Math & -- & -- & 0.86 & Base & -- & 1.00 \\
16 & Math & -- & -- & 1.00 & Base & -- & 0.99 \\
17 & Code & -- & -- & 1.00 & Math & -- & 1.09 \\
18 & Math & -- & -- & 1.00 & Math & -- & 1.00 \\
19 & Base & -- & -- & 1.00 & Base & -- & 0.90 \\
20 & Math & -- & -- & 1.00 & Math & -- & 1.26 \\
21 & Base & -- & -- & 1.00 & Base & -- & 1.24 \\
22 & Base & -- & -- & 1.00 & Base & -- & 1.00 \\
23 & Math & -- & -- & 0.96 & Math & -- & 0.82 \\
24 & Base & -- & -- & 0.85 & Math & -- & 1.08 \\
25 & Math & -- & -- & 1.01 & Base & -- & 1.00 \\
26 & Code & -- & -- & 1.00 & Base & -- & 1.00 \\
27 & Math & -- & -- & 1.00 & LM\_JA & -- & 1.00 \\
28 & Math & -- & -- & 1.00 & Math & -- & 1.07 \\
29 & Math & -- & -- & 1.00 & Base & -- & 1.00 \\
30 & Base & -- & -- & 1.00 & LM\_JA & Math & 0.87 \\
31 & Code & -- & -- & 1.04 & Math & LM\_JA & 0.84 \\
32 & Math & LM & -- & 0.89 & Math & LM\_JA & 0.99 \\
33 & Math & -- & -- & 1.00 & Math & -- & 1.00 \\
34 & Base & -- & -- & 1.00 & Math & LM\_JA & 0.73 \\
35 & LM & -- & -- & 1.00 & Math & -- & 1.00 \\
36 & Math & -- & -- & 0.98 & LM\_JA & -- & 1.00 \\
37 & Math & -- & -- & 1.00 & Math & -- & 1.00 \\
38 & Base & -- & -- & 1.00 & Math & LM\_JA & 1.00 \\
39 & Math & -- & -- & 0.95 & Math & -- & 1.00 \\
40 & Base & -- & -- & 1.00 & Math & -- & 1.00 \\
\hline
\end{tabular}
\label{tab:table12}
\end{table}

\end{document}